\documentclass{article}

\newif\ifCR

\usepackage{ijcai22}

\usepackage{times}

\usepackage{soul}
\usepackage{url}
\usepackage[hidelinks]{hyperref}
\usepackage[utf8]{inputenc}
\usepackage[small]{caption}
\usepackage{graphicx}
\usepackage{amsmath}
\usepackage{booktabs}
\usepackage[noend]{algpseudocode}
\usepackage{algorithm}
\usepackage{physics}
\usepackage{mathtools}
\usepackage{comment}
\usepackage{xcolor}
\usepackage{amssymb}
\usepackage{tabulary}

\newcommand{\appref}[1]{Appendix~\ref{#1}}
\newcommand{\sectref}[1]{Section~\ref{#1}}

\newcommand{\figref}[1]{Figure~\ref{#1}}
\newcommand{\tabref}[1]{Table~\ref{#1}}
\newcommand{\egref}[1]{Example~\ref{#1}}

\newcommand{\agref}[1]{Algorithm~\ref{#1}}

\newtheorem{example}{Example}

\newcommand{\blue}[1]{\textcolor{blue}{#1}}

\newcommand{\startpara}[1]{{\vskip1pt\noindent{\bf #1.}}} 
\renewcommand{\url}[1]{{\def~{\char126}\sf#1}}

\def\Nset{\mathbb{N}}

\def\cO{{\mathcal{O}}}

\def\cE{{\mathcal{E}}}
\def\cF{{\mathcal{F}}}
\def\cM{{\mathcal{M}}}
\def\cS{{\mathcal{S}}}
\def\cA{{\mathcal{A}}}
\def\cX{{\mathcal{X}}}

\def\cT{{\mathcal{T}}}
\def\cZ{{\mathcal{Z}}}

\newcommand{\rp}{\textit{resp. }}

\title{Toward Policy Explanations for Multi-Agent Reinforcement Learning}

\author{
 Kayla Boggess$^1$\and
 Sarit Kraus$^2$\And
 Lu Feng$^1$\\
 \affiliations
 $^1$University of Virginia\\
 $^2$Bar-Ilan University\\
 \emails
 \{kjb5we, lu.feng\}@virginia.edu, sarit@cs.biu.ac.il
 }

\begin{document}

\maketitle

\begin{abstract}
Advances in multi-agent reinforcement learning (MARL) enable sequential decision making for a range of exciting multi-agent applications such as cooperative AI and autonomous driving. Explaining agent decisions is crucial for improving system transparency, increasing user satisfaction, and facilitating human-agent collaboration. However, existing works on explainable reinforcement learning mostly focus on the single-agent setting and are not suitable for addressing challenges posed by multi-agent environments. We present novel methods to generate two types of policy explanations for MARL: (i) policy summarization about the agent cooperation and task sequence, and (ii) language explanations to answer queries about agent behavior. Experimental results on three MARL domains demonstrate the scalability of our methods. A user study shows that the generated explanations significantly improve user performance and increase subjective ratings on metrics such as user satisfaction.
\end{abstract}

\section{Introduction} \label{sec:intro} 

Recent years have witnessed a growing body of research in multi-agent reinforcement learning (MARL), enabling sequential decision making for a range of exciting multi-agent applications such as cooperative AI~\cite{dafoe2020open} and autonomous driving~\cite{kiran2021deep}.
Generating explanations about agent decisions is crucial for improving system transparency, increasing user satisfaction, and facilitating human-agent collaboration~\cite{kraus2020ai,chakraborti2020emerging}. 
However, existing works on explainable reinforcement learning (RL) mostly focus on the single-agent setting~\cite{wells2021explainable,heuillet2021explainability,puiutta2020explainable}.

Generating explanations for MARL agents that interact with each other in a common environment poses significant challenges.
The combinatorial nature of MARL (i.e., the joint state/action space grows exponentially with the number of agents) leads to scalability issues. 
Explanations should provide adequate information about agent behavior, including the interaction (e.g., cooperation) among multiple agents, for user understanding. 
Furthermore, explanations should avoid redundant information that may overwhelm or confuse users, thus decreasing user satisfaction and trust. 

To tackle these challenges, we develop novel methods to generate two types of policy explanations for MARL: (i) policy summarization, and (ii) query-based language explanations. 
Our methods rely on first building an abstract representation of MARL policy as a multi-agent Markov decision process (MMDP), which can be obtained by abstracting samples observed during the MARL policy evaluation.
The proposed method generates a summarization about the most probable sequence of agent behavior under a given MARL policy, by finding the most probable path through the MMDP abstraction and extracting information about agent cooperation and task sequence. 
The generated policy summarizations can help users to have a global view of agent decisions and support human-agent collaboration (e.g., users may adjust their workflow based on agents' task sequence). 

Additionally, we developed methods for generating language explanations to answer three types of queries about agent behavior, including ``When do [agents] do [actions]?'', ``Why don't [agents] do [actions] in [states]?'', ``What do [agents] do in [conditions]?''
Such explanations can enable users to understand specific agent decisions, debug faulty agent behavior, and refine user mental models.
Our work is inspired by the method proposed in~\cite{hayes2017improving}, which computes a minimal Boolean logic expression covering states satisfying the query criteria, and converts the Boolean expression to explanations via language templates. 
However, we find that a naive adaptation of this method for MARL generates explanations with redundant information and has limited scalability. 
Further, the generated explanations do not necessarily capture agent cooperation, which is imperative for explaining MARL. 
We proposed improved methods that address these limitations by leveraging MARL domain knowledge (e.g., agent cooperation requirements) to filter relevant agent states and actions.

We applied a prototype implementation of the proposed methods to three benchmark MARL domains: (i) multi-robot search and rescue, (ii) multi-robot warehouse, and (iii) level-based foraging~\cite{papoudakis2021benchmarking}.
Experimental results demonstrate that the proposed methods can generate policy summarizations and query-based explanations for large MARL environments with up to 19 agents.

Finally, we conducted a user study to evaluate the quality of generated explanations. 
We measured user performance on correctly answering questions based on explanations, to test user understanding of agent behavior. 
We also collected user subjective ratings on \emph{explanation goodness metrics}~\cite{hoffman2018metrics}. 
The results show that the generated explanations significantly improve user performance and increase subjective ratings on various metrics including user satisfaction.

\section{Related Work} \label{sec:related} 


Explaining agent decision making has recently emerged as a focus area within the explainable AI paradigm. 
\cite{chakraborti2020emerging} provides a survey about this emerging landscape of explainable decision making.
\cite{kraus2020ai} proposes Explainable Decisions in Multi-Agent Environments (xMASE) as a new research direction, emphasizing many challenges of generating multi-agent explanations, such as accounting for agent interactions and user satisfaction. 

Explainable RL has been attracting increasing interest, as shown in several recent surveys~\cite{wells2021explainable,heuillet2021explainability,puiutta2020explainable}. 
In particular, \cite{wells2021explainable} points out the lack of user studies as a major limitation 
across existing works. 
Moreover, current approaches mostly focus on the single-agent setting, 
while generating explanations for MARL has received scant attention so far. 
\cite{kazhdan2020marleme} extracts MARL model as abstract argumentations, which do not explicitly consider agent cooperation on the same tasks as in our work. 

The proposed methods are inspired by several prior works. 
\cite{topin2019generation} develops Abstracted Policy Graphs (i.e., Markov chains of abstract states) for explaining single-agent RL. Our work differs in that we adopt MMDP to represent multi-agent policy abstractions.
\cite{amir2018highlights} summarizes agent behavior by extracting trajectories from agent simulations and visualized them as videos.
Inspired by this idea, we created GIF animations as the baseline for evaluating policy summarizations in the user study. But we extracted the sequence of agent actions differently, focusing on the agent cooperation.
\cite{hayes2017improving} generates RL policy descriptions to answer queries. 
We adapted this method to multi-agent environments and proposed significant improvements, which will be described in \sectref{sec:query}.

\section{Policy Abstraction and Summarization} \label{sec:policy} 

We describe how to build an abstract representation of MARL policy in \sectref{sec:mmdp}, 
and present the method for generating policy summarization in \sectref{sec:sum}.

\subsection{Policy Abstraction}\label{sec:mmdp}

In the context of MARL, a group of $N$ agents interact with each other in a common environment and make decisions influenced by the joint states of all agents. 
Agent decisions can be captured by a joint policy $\pi: \cX \to \Delta (\cA)$,
which is a function mapping the set of joint states $\cX=\{(x^1, \dots, x^N)\}$ to a probabilistic distribution over the set of joint actions $\cA=\{(a^1, \dots, a^N)\}$, where $x^i$ (\rp $a^i$) denotes the state (\rp action) of agent $i$.
Once a policy is trained, agents can act upon it in any given state.
But there is a lack of a global view of the entire policy. 
Further, the size of the policy grows exponentially with the number of agents and state variables. 
To address these issues, we propose to build an abstract representation of the policy as the basis for generating explanations about agent behavior.

We use the multi-agent Markov decision process (MMDP) framework to represent MARL policy abstraction. 
Formally, an MMDP is a tuple $(\cS, \cA, \cT)$, 
where $\cS=\{(s^1, \dots, s^N)\}$ is the joint (abstract) state space,
$\cA$ is the joint action space, and $\cT$ is the transition function.
Let $\cF$ be a set of Boolean predicates indicating features of the MARL domain. 
We denote by $f(x^i)=1$ if an agent state $x^i$ satisfies a feature predicate $f \in \cF$.
An abstract state $s^i$ is then given by the satisfaction of all feature predicates $f \in \cF$, where each bit of the binary encoding of $s^i \in \Nset$ corresponds to the satisfaction of a predicate $f(x^i)$.
Thus, the choice of features affects the abstraction level and should include adequate information for explanations. 
In this work, we assume that users specify a set of feature predicates for a given MARL domain. 

Once an MARL policy is trained, we build an MMDP during the policy evaluation stage.
For each sample $(\vb{x},\vb{a},\vb{x}')$, 
determine an MMDP transition $\vb{s} \xrightarrow{\vb{a}}\vb{s}'$ 
by finding the abstract state $\vb{s}$ (\rp $\vb{s'}$) corresponding to $\vb{x}$ (\rp $\vb{x'}$). 
When policy evaluation terminates (e.g., converging to the expected reward),
compute the transition probability $\cT(\vb{s},\vb{a},\vb{s}')$ via frequency counting.

\startpara{Properties}
The resulting MMDP is a \emph{sound} abstraction of the MARL policy
because, by construction, every MMDP transition with non-zero probability corresponds to at least one sampled policy decision. 
The state space size $|\cS|$ is bounded by $\cO({2^{|\cF|}}^N)$, depending on the number of agents $N$ and feature predicates $|\cF|$.
In practice, a trained MARL policy may only induce a small set of reachable states.

\begin{figure}[t]
    \centering
    \includegraphics[width=.75\columnwidth]{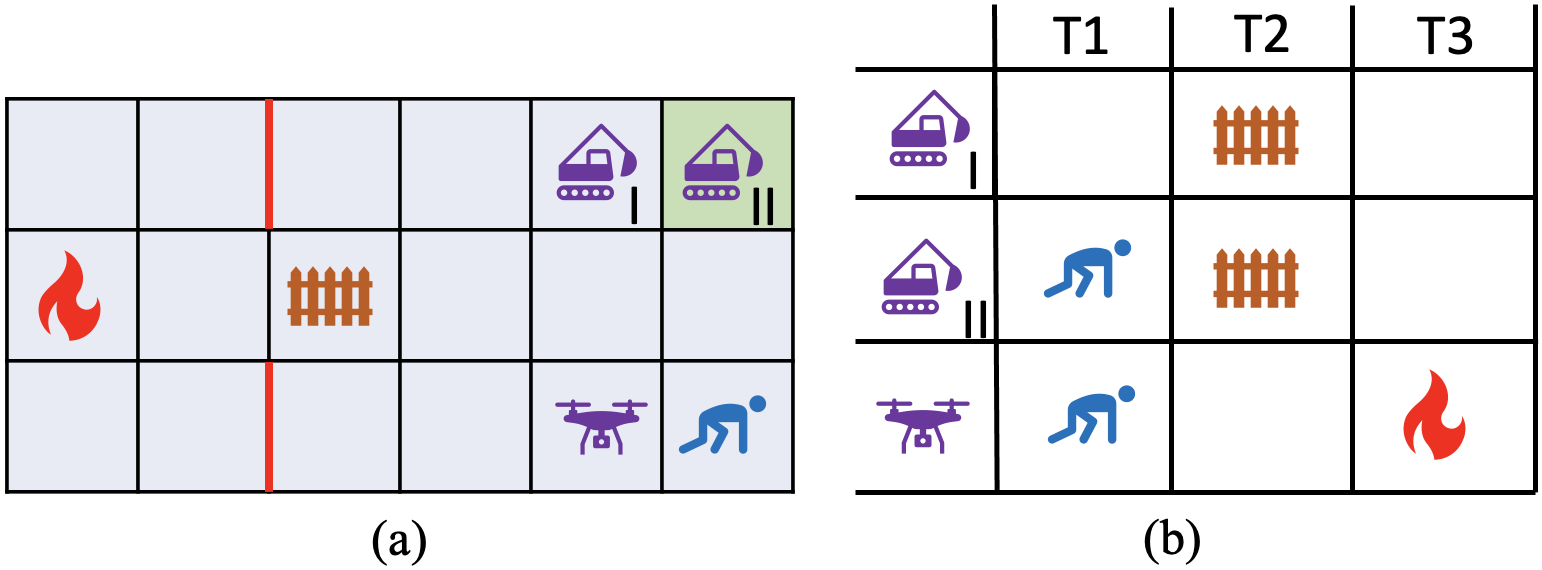}
    \caption{Example MARL domain of multi-robot search and rescue.}
    \vspace{-10pt}
    \label{fig:rs_eg}
\end{figure}

\begin{example} \label{eg:mmdp}
\figref{fig:rs_eg}(a) shows an example MARL domain where three robotic agents cooperate to complete search and rescue tasks. Rescuing the victim requires the cooperation of an unmanned aerial vehicle (UAV) and an unmanned ground vehicle (UGV). Any agent can fight the fire, which is blocked by the wall and obstacle. Removing the obstacle requires the cooperation of two UGVs. Given a trained MARL policy, we build an MMDP abstraction with 6 feature predicates indicating whether each task is detected or completed (e.g., $\mathsf{victim\_detect}$, $\mathsf{victim\_complete}$).
An agent can only detect a task in a neighboring grid (e.g., UAV detects the victim in \figref{fig:rs_eg}(a)). 
The resulting MMDP has 63 (reachable) states and 577 transitions.
\end{example}

\subsection{Policy Summarization}\label{sec:sum}

A policy abstraction containing hundreds of states and transitions is too complex for humans to understand.
An alternative way of communicating agent behavior is to show execution traces; however, a lengthy trace may be burdensome for users to review.
To overcome these limitations, we develop a method to generate policy summarization, illustrating the agent cooperation and task sequence for the most probable sequence of agent behavior under a given MARL policy.

\begin{algorithm}[tb]
\scriptsize
\caption{Generating Policy Summarization}
\label{ag:sum}
\textbf{Input}: policy abstraction $\cM=(\cS, \cA, \cT)$, task completion predicates $\cF_c \subseteq \cF$ \\
\textbf{Output}: policy summarization $\cZ$
\begin{algorithmic}[1] 
\State $\cZ \gets \{\}$
\State Compute the most probable path $\rho$ through $\cM$
\For{ $0 \le t \le |\rho|$}
    \State $y \gets $ new array
    \For{ $1 \le i \le N$}
        \State $y[i] \gets$ \{\}
        \For{$f \in \cF_c$}
            \If{agent state $s^i_t$ in the path $\rho$ satisfies $f$}
                \State insert $f$ to $y[i]$
            \EndIf
        \EndFor
    \EndFor
    \State insert non-empty array $y$ to $\cZ$
\EndFor
\State \Return $\cZ$
\end{algorithmic}
\end{algorithm}


\agref{ag:sum} shows the proposed method, which takes the input of a policy abstraction $\cM$ and a set of predicates $\cF_c$ representing the completion of tasks (subgoals) in a given MARL domain. 
The first step is to compute the most probable path $\rho = \vb{s}_0 \xrightarrow{\vb{a}_0} \vb{s}_1 \xrightarrow{\vb{a}_1} \cdots$ from the initial state to a goal state in the MMDP $\cM$,
which represents the most probable sequence of agent decisions under the policy. 
This problem can be solved by converting the MMDP to a directed weighted graph with edge weight $e(\vb{s},\vb{a},\vb{s'}) = -\log \cT(\vb{s},\vb{a},\vb{s'})$ for each transition,
and then applying the Dijkstra's algorithm~\cite{dijkstra1959note} to find the shortest path. 

Next, the algorithm loops through every joint state $\vb{s}_t$ in the path $\rho$ to extract the agent cooperation and task sequence.
At each step $t$, the algorithm checks if an agent state $s^i_t$ satisfies any task completion predicate $f \in \cF_c$
and inserts completed task $f$ into the array element $y[i]$ (line 4-9).
An agent only satisfies a task completion predicate at step $t$ when it finishes the task and receives a reward.
We assume that if a task is completed via the cooperation of multiple agents, 
they must satisfy the task predicate $f$ at the same step $t$ and each receive a portion of the reward.
Thus, the agent cooperation is represented as multiple elements of the array $y$ sharing the same task. 
Only non-empty arrays containing completed tasks are inserted into the summarization $\cZ$.
When the algorithm terminates, the generated summarization is visualized as a chart,
with each column corresponding to a non-empty $y$-array and each row representing an agent's task sequence.

\startpara{Properties}
The generated policy summarization $\cZ$ is \emph{sound}, because it is derived from the most probable path of a \emph{sound} policy abstraction (see \sectref{sec:mmdp}). 
The complexity of computing the most probable path is bounded by $\cO(|\cS|^2)$, following the complexity of the Dijkstra's algorithm and depending on the MMDP state space size.  
The rest of \agref{ag:sum} is bounded by $\cO(|\rho| \cdot N \cdot |\cF_c|)$, depending on the path length and the number of agents and tasks.

\begin{example} \label{eg:sum}
We apply \agref{ag:sum} using the policy abstraction and task predicates from \egref{eg:mmdp}.
There are 8 states in the most probable path from the initial state (i.e., all agents starting in the green grid) to a goal state (i.e., all tasks have been completed). 
\figref{fig:rs_eg}(b) visualizes the generated summarization, with column names (i.e., T1, T2, T3) indicating the sequence of task completions:
UGV$_2$ and UAV cooperate to rescue the victim; next, UGV$_1$ and UGV$_2$ cooperate to remove the obstacle; and lastly, UAV fights the fire. 
\end{example}

\section{Query-Based Explanations} \label{sec:query} 

While policy summarization provides a global view of the agent behavior under a MARL policy, 
users may also query about specific agent decisions. 
In this section, we develop methods for generating language explanations to answer the following three types of queries:
\begin{itemize}
    \item \emph{``When do [agents] do [actions]?''} for identifying conditions for action(s) of a single or multiple agent(s) .
    \item \emph{``Why don't [agents] do [actions] in [states]?''} for understanding differences in expected and observed behaviors of a single or multiple agent(s).
    \item \emph{``What do [agents] do in [predicates]?''} for revealing agent behavior under specific conditions described by the given predicates. 
\end{itemize}
Our work is inspired by a method developed in~\cite{hayes2017improving} to generate query-based explanations for single-agent RL. 
In the following, we propose new methods to tackle limitations posed by adapting this baseline method to multi-agent environments.

\subsection{Explanations for When Query}\label{sec:when}

\agref{ag:when} presents both the baseline and proposed methods for answering ``When do agents $G_q$ do actions $A_q$?'', where $G_q$ and $A_q$ are sets of agents and actions, respectively. The text in blue highlights changes about relevancy filters (RF) for the proposed method (called WithRF) compared to the baseline (called NoRF).

WithRF starts the algorithm (line 1-5) by identifying relevant agents $G$, features $F$, and action sets $A$ based on domain knowledge (e.g., agent cooperation requirements). 
For example, consider a query ``When does UAV rescue the victim?''. 
The domain knowledge is that rescuing the victim requires the cooperation of a UAV and a UGV (\egref{eg:mmdp}). 
Thus, the relevant agent set $G$ is \{$\mathsf{UVA, UGV\_1, UGV\_2}$\}.
The relevant feature set $F$ is \{$\mathsf{victim\_detect, victim\_complete}$\}, while predicates about the fire and obstacle are irrelevant. 
The relevant action sets $A$ is an array with each element representing one possible set of agent actions required for cooperation: 
$[\{\mathsf{UAV\_rescue}, \mathsf{UGV_1\_rescue}\}, \{\mathsf{UAV\_rescue}, \mathsf{UGV_2\_rescue}\}]$,
which can be generated based on the aforementioned domain knowledge about agent cooperation requirements. 

Both NoRF and WithRF loop through all the joint states $\vb{s} \in \cS$ of the policy abstraction MMDP and 
check all the \emph{enabled} (i.e., with non-zero transition probability) joint actions $\vb{a}$ in state $\vb{s}$.
In line 9 of \agref{ag:when}, NoRF checks if $\vb{a}$ is \emph{compatible} with $A_q$; 
that is, every agent action $a \in A_q$ is contained in the joint action $\vb{a}=(a^1, \dots, a^N)$.
By contrast, WithRF checks if $\vb{a}$ is compatible with at least one set of relevant actions contained in the array $A$. 
Following the previous example, NoRF checks if $\vb{a}$ contains $\mathsf{UAV\_rescue}$,
while WithRF checks if $\vb{a}$ contains $\{\mathsf{UAV\_rescue}, \mathsf{UGV_1\_rescue}\}$ \emph{or} $\{\mathsf{UAV\_rescue}, \mathsf{UGV_2\_rescue}\}$. 
Since each element of $A$ is a super-set of $A_q$, the WithRF check is more restrictive.

\begin{algorithm}[tb]
\scriptsize
\caption{Generating Query-Based Explanations}
\label{ag:when}
\textbf{Input}: policy abstraction $(\cS, \cA, \cT)$, 
    query ``when do agents $G_q$ do actions $A_q$?'' \\
\textbf{Output}: explanations $\cE$
\begin{algorithmic}[1] 
\State \blue{$G \gets \{\}$; $F \gets \{\}$; $A \gets [\{\}]$} 
\ForAll{\blue{agent action $a^i \in A_q$}}
    \State \blue{insert all relevant agents of $a^i$ to $G$}
    \State \blue{insert all relevant features of $a^i$ to $F$}
    \State \blue{insert all relevant action sets of $a^i$ to $A$}
\EndFor
\State $V \gets \{\}$; $\bar{V} \gets \{\}$
\ForAll{joint state $\vb{s} \in \cS$}
    \ForAll{joint action $\vb{a}$ enabled in $\vb{s}$}
        \If{$\vb{a}$ is compatible with $A_q$ \blue{[replace $A_q$  with $A$]}}
            \State insert $\vb{s}$ to $V$
        \Else
            \State insert $\vb{s}$ to $\bar{V}$
        \EndIf
    \EndFor    
\EndFor 
\State $B_1 \gets$ States2Boolean($V$); $B_0 \gets$ States2Boolean($\bar{V}$)
\State $\phi \gets$ Quine-McCluskey(ones=$B_1$, zeros=$B_0$) 
\State translate $\phi$ to explanations $\cE$ via language templates
\State \Return $\cE$
\Function{State2Boolean}{$W$}
\State $B \gets \{\}$
\ForAll{$\vb{s}=(s^1, \dots, s^N) \in W$}
    \For{$1 \le i \le N$ \blue{[replace with $i \in G$]}}
        \State $C \gets$ feature predicate valuations of state $s^i$
        \For{$f \in \cF$ \blue{[replace with $f \in F$]}}
            \State insert $C(f)$ to $B$
        \EndFor
    \EndFor
\EndFor
\State \Return $B$
\EndFunction
\end{algorithmic}
\end{algorithm}

If a state $\vb{s}$ has at least one enabled action $\vb{a}$ passing the aforementioned checks, $\vb{s}$ is inserted to the target states set $V$; and to the non-target states set $\bar{V}$ otherwise.
The intuition is that the generated explanations should describe target states satisfying the query criteria and exclude conditions of non-target states.
WithRF poses further restrictions that target states need to satisfy criteria captured by relevant actions $A$, such as agent cooperation requirements.
Thus, explanations generated by WithRF can provide information about agent cooperation, which may be missed by NoRF explanations. 

Next, the algorithm converts the states set $V$ (\rp $\bar{V}$) to a list of Boolean formulas $B_1$ (\rp $B_0$) via the function described in line 17-24.
Given a joint state $\vb{s}=(s^1, \dots, s^N)$, 
NoRF finds valuations of every feature predicates $f \in \cF$ for all agent state $s^i$ and insert them to the list $B$.
By contrast, WithRF only inserts to $B$ the valuations of relevant features $f \in F$ in relevant agent states $s^i$ for all $i \in G$.
Following the previous example, WithRF only considers Boolean formulas related to relevant feature predicates 
\{$\mathsf{victim\_detect, victim\_complete}$\}, filtering out features related to the fire and obstacle.

Lastly, the algorithm supplies Boolean formulas $B_1$ and $B_0$ to the Quine-McCluskey algorithm~\cite{quine1952problem} and obtains a minimized Boolean formula, which can be translated into language explanations following~\cite{hayes2017improving}.
The runtime of Quine-McCluskey grows exponentially with the number of variables. 
Thus, WithRF is more efficient than NoRF, due to the decreased number of Boolean variables.
Moreover, filtering out irrelevant agents and features helps WithRF to prevent redundant information in the generated explanations.

\startpara{Properties}
Following the Quine-McCluskey, the complexity of NoRF is bounded by $\cO\big(3^{N \cdot |\cF|} / \ln( N \cdot |\cF|)\big)$.
The complexity of WithRF is reduced to 
$\cO\big(3^{|G|\cdot |F|} / \ln(|G|\cdot |F|)\big)$.

\begin{table*}[t]
\tiny
\centering
\begin{tabulary}{\textwidth}{L|L|L}
\toprule
\textbf{Query}  & \textbf{Explanations generated by NoRF (baseline)} & \textbf{Explanations generated by WithRF (proposed)} \\
\midrule
When does UAV rescue the victim?
& UAV rescues the victim when UAV detects the victim \emph{and} UGV$_1$ does not detect the fire, \emph{or} UAV detects the victim \emph{and} UGV$_1$ does not detect the obstacle.
& UAV rescues the victim when UAV detects the victim \emph{and} UGV$_1$ detects the victim, \emph{or} UAV detects the victim \emph{and} UGV$_2$ detects the victim. 
\\ \midrule
Why don't UGV$_1$ and UGV$_2$ remove the obstacle in this state?
& UGV$_1$ and UGV$_2$ don't remove the obstacle in this state because UGV$_1$ does not detect the obstacle. 
& UGV$_1$ and UGV$_2$ don't remove the obstacle in this state because UGV$_1$ does not detect the obstacle \emph{and} UGV$_2$ does not detect the obstacle. 
\\ \midrule
What does UAV do when it detects the victim?
& UAV can rescue the victim, move, or wait when it detects the victim.
& UAV is most likely to rescue the victim when it detects the victim.
\\ \bottomrule
\end{tabulary}
\caption{Examples of query-based explanations}
\label{tab:qe}
\end{table*}

\begin{example} \label{eg:when}
\tabref{tab:qe} (first row) shows the explanations generated by NoRF and WithRF for a when query. 
The NoRF explanation contains redundant information about the fire and obstacle that are irrelevant to the query. 
The WithRF explanation completely captures the required agent cooperation for the query, which is missed by the NoRF explanation. 
\end{example}

\subsection{Explanations for Why Not Query }\label{sec:why}

The query ``Why don't agents $G_q$ do actions $A_q$ in the joint State $\vb{s}_q$?''
can be answered by modifying \agref{ag:when} as follows. 
In line 10, adding $\vb{s}$ to $\bar{V}$ instead of $V$. 
Remove line 11-12 and add a new line for inserting the query state $\vb{s}_q$ to $V$.
The modified algorithm 
\ifCR 
(see Appendix A in~\cite{boggess2022marl}) 
\else 
(see \appref{app:ag}) 
\fi
generates an explanation describing the differences between the observed behavior in the target query state $\vb{s}_q$ and the expected behavior of states with actions compatible with the query actions $A_q$ (NoRF) or relevant action sets $A$ (WithRF). 
The complexity of the modified algorithm follows \agref{ag:when}.

\begin{example} \label{eg:why}
\tabref{tab:qe} (second row) shows the explanations generated by NoRF and WithRF for a why not query about the behavior of two agents UGV$_1$ and UGV$_2$ in the state shown in \figref{fig:rs_eg}(a).
The WithRF explanation captures the required agent cooperation for removing the obstacle, while the NoRF explanation fails to provide such information. 
\end{example}

\subsection{Explanations for What Query }\label{sec:what}

To answer the query ``What do agents $G_q$ do when satisfying predicates $F_q$?'', 
we first identify all the satisfying joint states $\vb{s}=(s^1, \dots, s^N)$;
that is, for all $i \in G_q$, agent state $s^i$ satisfies predicates $F_q$.
The baseline NoRF method is to generate a list of all possible enabled actions for agents $G_q$ in these states. 
The proposed WithRF method improves the baseline by filtering agent actions that are relevant to predicates $F_q$ and finding the most likely relevant agent actions from the list via frequency counting. 
Since the proposed methods\footnote{See 
\ifCR 
Appendix A in~\cite{boggess2022marl}
\else 
\appref{app:ag} 
\fi
for the pseudos code. } 
do not need to call the Quine-McCluskey, the complexity of both NoRF and WithRF are only bounded by $\cO(|G_q| \cdot |\cS| \cdot |\cA|)$, depending on the number of query agents, joint state space, and joint action space of the policy abstraction.

\begin{example} \label{eg:what}
\tabref{tab:qe} (third row) shows the explanations generated by NoRF and WithRF for a what query. 
The WithRF explanation is more concise than the NoRF explanation and only contains relevant action for the query predicate. 
\end{example}

\section{Computational Experiments} \label{sec:eval} 

\begin{table*}[t]
\resizebox{\textwidth}{!}{%
\begin{tabular}{ccllccllclcllcclccllcclccllcclcc}
\toprule
\multicolumn{2}{c}{\textbf{Case Study}} &  &  & \multicolumn{2}{c}{\textbf{MMDP}} &  &  & \multicolumn{3}{c}{\textbf{Summarization}} &  &  & \multicolumn{5}{c}{\textbf{When Query}} &  &  & \multicolumn{5}{c}{\textbf{Why Not Query}} &  &  & \multicolumn{5}{c}{\textbf{What Query}} \\ \cline{1-2} \cline{5-6} \cline{9-11} \cline{14-18} \cline{21-25} \cline{28-32} 
 &  &  &  &  &  &  &  & \textbf{Path} &  & \textbf{Chart} &  &  & \multicolumn{2}{c}{\textbf{NoRF}} &  & \multicolumn{2}{c}{\textbf{WithRF}} &  &  & \multicolumn{2}{c}{\textbf{NoRF}} &  & \multicolumn{2}{c}{\textbf{WithRF}} &  &  & \multicolumn{2}{c}{\textbf{NoRF}} &  & \multicolumn{2}{c}{\textbf{WithRF}} \\ \cline{9-9} \cline{11-11} \cline{14-15} \cline{17-18} \cline{21-22} \cline{24-25} \cline{28-29} \cline{31-32} 
Domain & $N$ &  &  & $|\cS|$ & $|\cT|$ &  &  & $|\rho|$ &  & $|\cZ|$ &  &  & $|\cE|$ & Time (ms) &  & $|\cE|$ & Time (ms) &  &  & $|\cE|$ & Time (ms) &  & $|\cE|$ & Time (ms) &  &  & $|\cE|$ & Time (ms) &  & $|\cE|$ & Time (ms) \\ \midrule
 & 3 & \multicolumn{1}{l|}{} &  & 63 & 577 & \multicolumn{1}{l|}{} &  & 8 &  & 3x2 & \multicolumn{1}{l|}{} &  & 4 & 72.4 &  & 4 & 1.2 & \multicolumn{1}{l|}{} &  & 1 & 86.8 &  & 2 & 0.8 & \multicolumn{1}{l|}{} &  & 3 & 1.7 &  & 1 & 1.3 \\
SR & 4 & \multicolumn{1}{l|}{} &  & 732 & 5,048 & \multicolumn{1}{l|}{} &  & 23 &  & 4x4 & \multicolumn{1}{l|}{} &  & - & timeout &  & 6 & 31.4 & \multicolumn{1}{l|}{} &  & - & timeout &  & 3 & 14.9 & \multicolumn{1}{l|}{} &  & 3 & 15.7 &  & 1 & 14.5 \\
 & 5 & \multicolumn{1}{l|}{} &  & 839 & 4,985 & \multicolumn{1}{l|}{} &  & 24 &  & 5x4 & \multicolumn{1}{l|}{} &  & - & timeout &  & 10 & 73.6 & \multicolumn{1}{l|}{} &  & - & timeout &  & 4 & 23.4 & \multicolumn{1}{l|}{} &  & 6 & 24.6 &  & 1 & 19.8 \\ \midrule
 & 2 & \multicolumn{1}{l|}{} &  & 8 & 62 & \multicolumn{1}{l|}{} &  & 5 &  & 2x2 & \multicolumn{1}{l|}{} &  & 4 & 2.6 &  & 4 & 2.7 & \multicolumn{1}{l|}{} &  & 4 & 2.0 &  & 2 & 0.4 & \multicolumn{1}{l|}{} &  & 3 & 0.1 &  & 1 & 0.1 \\
RWARE & 4 & \multicolumn{1}{l|}{} &  & 16 & 387 & \multicolumn{1}{l|}{} &  & 5 &  & 4x1 & \multicolumn{1}{l|}{} &  & 12 & 3,321.0 &  & 5 & 82.7 & \multicolumn{1}{l|}{} &  & 6 & 23.0 &  & 3 & 1.3 & \multicolumn{1}{l|}{} &  & 3 & 0.1 &  & 1 & 0.1 \\
 & 19 & \multicolumn{1}{l|}{} &  & 114 & 1,500 & \multicolumn{1}{l|}{} &  & 15 &  & 19x2 & \multicolumn{1}{l|}{} &  & - & timeout &  & 7 & 3,890.6 & \multicolumn{1}{l|}{} &  & - & timeout &  & 4 & 56.5 & \multicolumn{1}{l|}{} &  & 3 & 21.7 &  & 1 & 20.1 \\ \midrule
 & 2 & \multicolumn{1}{l|}{} &  & 5 & 13 & \multicolumn{1}{l|}{} &  & 4 &  & 2x2 & \multicolumn{1}{l|}{} &  & 2 & 0.7 &  & 2 & 0.7 & \multicolumn{1}{l|}{} &  & 3 & 0.4 &  & 2 & 0.3 & \multicolumn{1}{l|}{} &  & 2 & 0.1 &  & 1 & 0.1 \\
LBF & 4 & \multicolumn{1}{l|}{} &  & 15 & 355 & \multicolumn{1}{l|}{} &  & 5 &  & 4x1 & \multicolumn{1}{l|}{} &  & 10 & 3,598.5 &  & 2 & 1.4 & \multicolumn{1}{l|}{} &  & 8 & 3,695.0 &  & 2 & 1.6 & \multicolumn{1}{l|}{} &  & 3 & 0.6 &  & 1 & 0.6 \\
 & 9 & \multicolumn{1}{l|}{} &  & 482 & 5,841 & \multicolumn{1}{l|}{} &  & 13 &  & 9x2 & \multicolumn{1}{l|}{} &  & - & timeout &  & 2 & 200.2 & \multicolumn{1}{l|}{} &  & - & timeout &  & 2 & 101.3 & \multicolumn{1}{l|}{} &  & 2 & 24.9 &  & 1 & 19.8 \\ \bottomrule
\end{tabular}%
}
\caption{Experimental results on three MARL domains (timeout set as one hour).}
\vspace{-10pt}
\label{tab:exp}
\end{table*}

We implemented and applied the proposed methods to three MARL domains.
The first domain is multi-robot search and rescue (SR) similar to \egref{eg:mmdp}. 
The second and third domains are benchmarks taken from~\cite{papoudakis2021benchmarking}.
Multi-robot warehouse (RWARE) considers multiple robotic agents cooperatively delivering requested items.
Level-based foraging (LBF) considers a mixed cooperative-competitive game where agents must navigate a grid world to collect randomly scattered food.
Our implementation used the Shared Experience Actor-Critic~\cite{christianos2020shared} for MARL policy training and evaluation.
All models were trained and evaluated to 10,000 steps, or until converging to the expected reward, whichever occurred first. 
The experiments were run on a laptop with a 1.4 GHz Quad-Core Intel i5 processor and 8 GB RAM.

\tabref{tab:exp} shows experimental results. 
For each domain, we report the number of agents $N$ and the number of states $|\cS|$ and transitions $|\cT|$ of generated policy abstraction MMDP. It is unsurprising that the MMDP size grows exponentially with the number of agents.
We report the most probable path length $|\rho|$ and the chart size $|\cZ|$ of generated policy summarizations,
which are more compact and easier to interpret than complex MMDP abstractions.
All summarizations were generated within 1 second (thus not shown in the table).
Additionally, we compare NoRF and WithRF methods in terms of the number of clauses in the generated query-based explanations and runtime. 
The results show that WithRF is more succinct in general and has better scalability than NoRF. 
In particular, NoRF failed to generate explanations for ``when'' and ``why not'' queries within an hour for large cases with more than 4 agents, while WithRF generated explanations for all cases within seconds. 
Both methods generate explanations for ``what'' queries efficiently, thanks to the lower complexity than other queries (see \sectref{sec:query}).

In summary, experimental results demonstrate that the proposed methods can generate policy summarizations and query-based explanations for large MARL domains (e.g., RWARE with 19 agents, which is the largest number of possible agents in the provided environments).

\section{User Study} \label{sec:exp} 

We conducted a user study~\footnote{This study was approved by University of Virginia Institutional Review Boards IRB-SBS \#4701. 
Study details (e.g., questionnaire, interface) are included in 
\ifCR 
Appendix B of~\cite{boggess2022marl}.
\else 
\appref{app:study}.
\fi
}
via Qualtrics to evaluate the quality of generated explanations. We describe the study design in \sectref{sec:study_design} and analyze results in \sectref{sec:study_results}. 

\subsection{Study Design} \label{sec:study_design}

\startpara{Participants} 
We recruited 116 eligible participants (i.e., fluent English speakers over the age of 18) through university mailing lists. 
62.1\% of participants self-identified as male, 37.1\% as female, and 0.8\% preferred not to say. 
The age distribution is 76(18-24), 31(25-34), 7(35-49), 2(50-64).
Participants were instructed to answer multiple-choice questions about agent behavior for multi-robot search and rescue tasks.
They were incentivized with bonus payments to answer questions correctly based on the provided explanations. 
To ensure data quality, attention checks were injected during the study.

\startpara{Independent variables}
We employed a within-subject study design with the explanation generation methods as independent variables. 
Participants were asked to complete two trials for evaluating policy summarizations.
They were presented with charts generated by \agref{ag:sum} in one trial, 
and GIF animations illustrating the most probable sequence of agent behavior (i.e., visualization of the most probable path in the policy abstraction) in the other trial. 
For each trial, there were two questions about agent behavior in various environments (i.e., 3$\times$6 and 6$\times$6 grid world). Questions used in the two trials are different but had similar difficulty. 
All participants were presented with the same set of four randomly generated questions for summarization trials. 
To counterbalance the ordering confound effect, they were randomly assigned to answer the first two questions based on either charts or GIF, and the other two questions based on the remaining method. 
Additionally, participants were asked to complete two trials for evaluating query-based explanations generated by NoRF and WithRF methods, with 6 questions (2 environments $\times$ 3 query types) in each trial. 
Participants answered the same set of 12 randomly generated questions for query-based trials, and were randomly assigned to different groups similarly to summarization trials.

\startpara{Dependent measures} 
We measured \emph{user performance} by counting the number of correctly answered questions in each trial. 
In addition, at the end of each trial, participants were asked to rate in a 5-point Likert scale (1 - strongly disagree, 5 - strongly agree) about \emph{explanation goodness} metrics (i.e., understanding, satisfaction, detail, completeness, actionability, reliability, trust)~\cite{hoffman2018metrics}.

\startpara{Hypotheses}
We make the following hypotheses in this study.
\begin{itemize}
    \item \textbf{H1:} Chart-based summarizations lead to better user performance than GIF-based.
    \item \textbf{H2:} Chart-based summarizations yield higher user ratings on explanation goodness metrics than GIF-based.
    \item \textbf{H3:} Query-based explanations generated by WithRF lead to better user performance than those by NoRF. 
    \item \textbf{H4:} Query-based explanations generated by WithRF yield higher user ratings on explanation goodness metrics than those by NoRF. 
\end{itemize}

\subsection{Results Analysis} \label{sec:study_results}

\begin{figure}[t]
    \centering
    \includegraphics[width=.85\columnwidth]{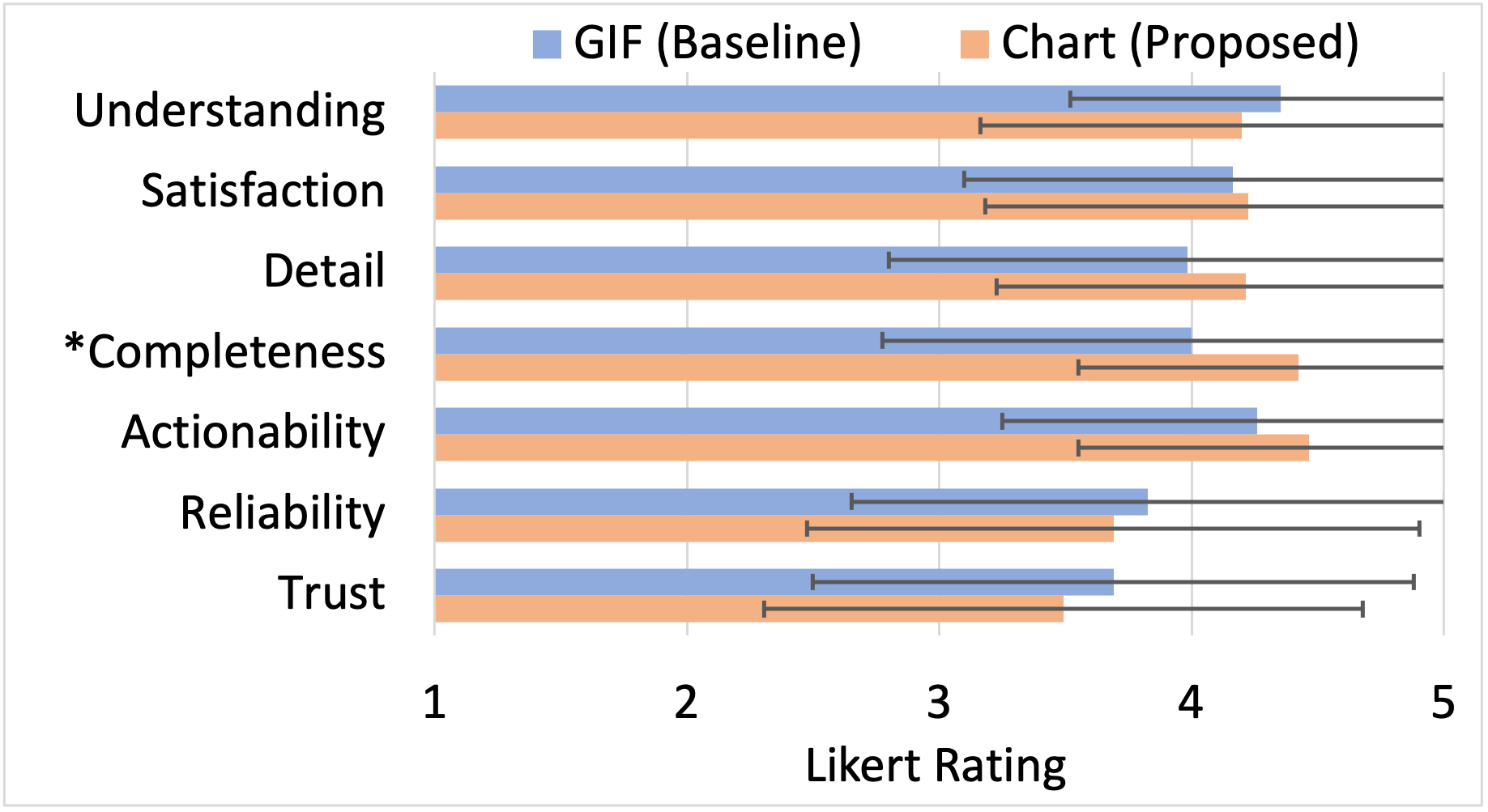}
    \caption{Mean and SD of participant ratings about policy summarizations (``*'' indicates statistically significant difference).}
    \label{fig:sumGood}
\end{figure}

We used a paired t-test to evaluate hypotheses H1 and H3, and used the Wilcoxon Signed-rank test to evaluate hypotheses H2 and H4. We set the significant level as $\alpha=0.05$.

\startpara{Evaluating policy summarizations}
Participants answered more questions correctly with chart-based summarizations (M=1.8 out of 2, SD=0.6) than GIF-based (M=0.9 out of 2, SD=0.4), 
with statistical significance ($t$(462)=-15.8, $p\le$0.01, $d$=1.5).
\emph{Thus, the data supports H1.}

\figref{fig:sumGood} shows average participant ratings about summarizations.
Chart-based summarizations yield higher ratings on the perceived completeness than GIF-based with statistical significance ($W$=371.5, $Z$=-2.4, $p \leq$0.02, $r$=-0.2).
But no significant difference was found regarding the other metrics.
\emph{Thus, the data partially supports H2.}

\startpara{Evaluating query-based explanations}
Participants answered more questions correctly with explanations generated by WithRF (M=5.2 out of 6, SD=1.7) than NoRF (M=2.3 out of 6, SD=1.0), with statistical significance ($t$(1390)=-21.1, $p\le$0.01,$d$=2.0).
\emph{Thus, the data supports H3.}

\begin{figure}[t]
    \centering
    \includegraphics[width=.85\columnwidth]{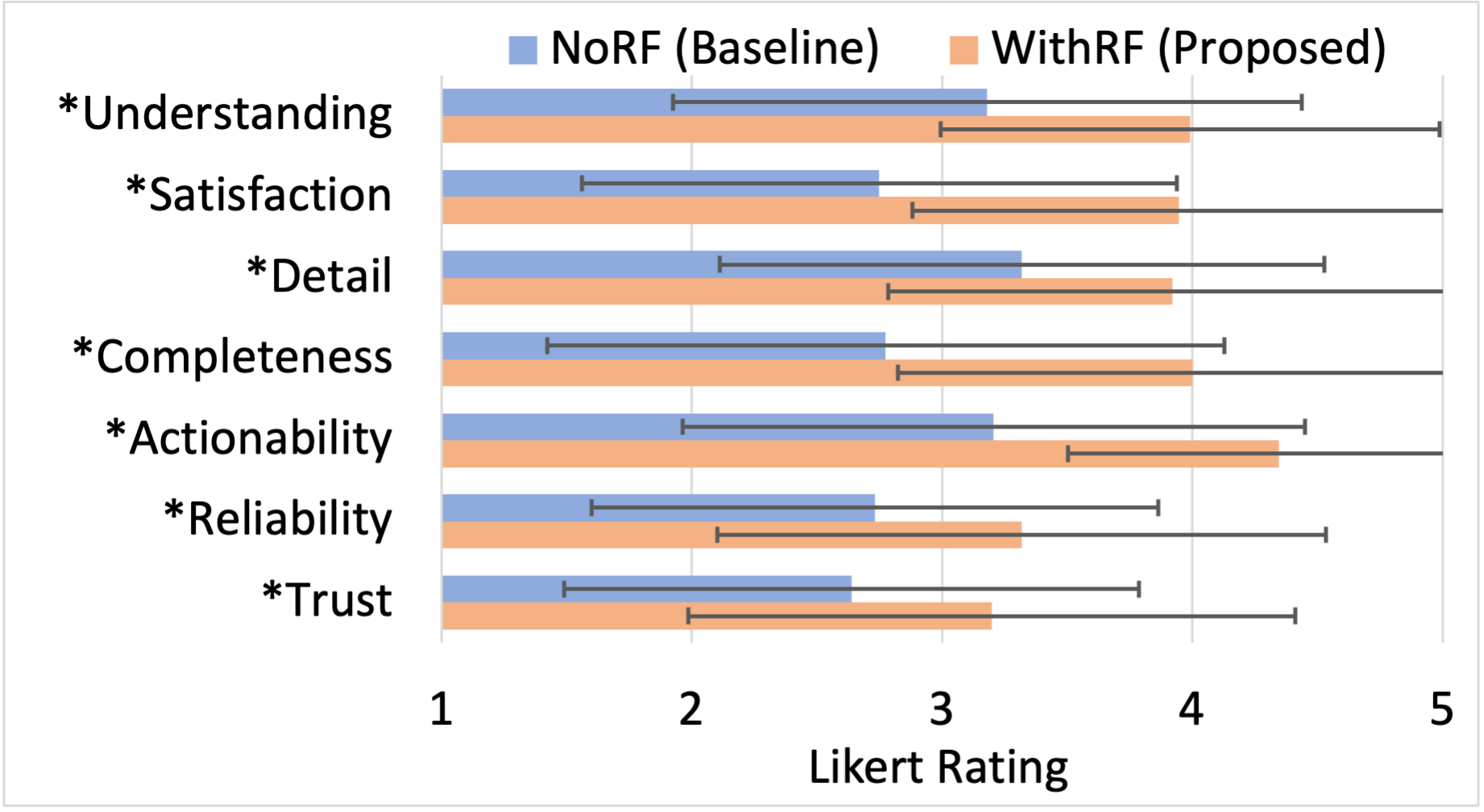}
    \caption{Mean and SD of participant ratings about query-based explanations (``*'' indicates statistically significant difference).}
    \label{fig:queryGood}
\end{figure}

\figref{fig:queryGood} shows that participants gave higher average ratings to WithRF explanations than NoRF explanations. 
The Wilcoxon test found significant differences on all metrics: understanding ($W$=319.5, $Z$=-4.9, $p \leq$0.01, $r$=-0.3), satisfaction ($W$=266.0, $Z$=-7.0, $p \leq$0.01, $r$=-0.5), detail ($W$=484.0, $Z$=-3.7, $p \leq$0.01, $r$=-0.2), completeness ($W$=494.5, $Z$=-6.4, $p \leq$0.01, $r$=-0.4), actionability ($W$=167.0, $Z$=-6.9, $p \leq$0.01, $r$=-0.5), reliability ($W$=382.5, $Z$=-3.6, $p \leq$0.01, $r$=-0.2), and trust ($W$=217.0, $Z$=-3.4, $p \leq$0.01, $r$=-0.2). 
\emph{Thus, the data supports H4.}

\startpara{Discussion}
In summary, the data supports all hypotheses, while H2 is only partially supported because the statistical test found no significant differences between chart-based and GIF-based summarizations on most metrics.
However, \figref{fig:sumGood} shows that participants rated chart-based summarizations close to 4 (agree) on all metrics, and above GIF-based ratings on all metrics except understanding, reliability, and trust.
This may be because users showed a strong preference toward the moving nature of GIF animations and the visualized effects of agents completing tasks.
But watching a GIF can be more time-consuming and less informative than a quick glance at the chart. 
This is supported by the results that participants were able to answer more questions correctly with chart-based summarizations, and they rated this method significantly higher on completeness (i.e., providing needed information). 
Meanwhile, query-based explanations generated by the proposed WithRF method led to significantly better user performance and higher user ratings on all metrics,
because users prefer succinct WithRF explanations with adequate information about agent behavior and cooperation.
By contrast, NoRF explanations do not necessarily provide essential information about agent cooperation for correctly answering questions, and may contain redundant information that decreases user satisfaction.

\section{Conclusion} \label{sec:conclu} 

In this work, we developed methods to generate policy summarizations and query-based explanations for MARL.
Experimental results on three MARL domains demonstrate the scalability of our methods. 
Evaluation via a user study shows that our generated MARL policy explanations can improve user understanding about agent behavior and enable them to answer more questions correctly, while maintaining  very positive ratings on explanation goodness metrics.

The proposed methods are independent from the inner working of MARL methods, only relying on policy abstractions that can be built via observing samples during the policy evaluation. Although we only used one MARL method in our experiments, the proposed methods can be used to generate policy explanations for different MARL methods. 
As part of the future work, we plan to apply the proposed methods to a wide range of MARL methods and domains.

\clearpage
\section*{Acknowledgments}
This work was supported in part by U.S. National Science Foundation grant CCF-1942836, U.S. Office of Naval Research grant N00014-18-1-2829, Israel Science Foundation grant 1958/20, EU Project TAILOR grant 952215, and by the Data Science Institute at Bar-Ilan
University.
Any opinions, findings, and conclusions or recommendations expressed in this material are those of the author(s) and do not necessarily reflect the views of the grant sponsors.

\bibliographystyle{named}
\bibliography{references}

\ifCR 
\else 
\clearpage
\appendix

\section{Algorithms} \label{app:ag}
The text in blue highlight changes about relevancy filters (RF) 
for the proposed WithRF method compared to the baseline NoRF method.

\begin{algorithm}[h]
\footnotesize
\caption{Answering ``why not'' query}
\label{ag:whynot}
\textbf{Input}: policy abstraction $\cM=(\cS, \cA, \cT)$, query ``why don't agents $G_q$ do actions $A_q$ in the joint State $\vb{s}_q$''  \\
\textbf{Output}: language explanations $\cE$
\begin{algorithmic}[1] 
\State \blue{$G \gets \{\}$; $F \gets \{\}$; $A \gets [\{\}]$} 
\ForAll{\blue{agent action $a^i \in A_q$}}
    \State \blue{insert all relevant agents of $a^i$ to $G$}
    \State \blue{insert all relevant features of $a^i$ to $F$}
    \State \blue{insert all relevant action sets of $a^i$ to $A$}
\EndFor
\State $V \gets \{\}$; $\bar{V} \gets \{\}$
\State insert $\vb{s}_q$ to $V$
\ForAll{joint state $\vb{s} \in \cS$}
    \ForAll{joint action $\vb{a}$ enabled in $\vb{s}$}
        \If{$\vb{a}$ is compatible with $A_q$ \blue{[replace $A_q$  with $A$]}}
            \State insert $\vb{s}$ to $\bar{V}$
        \EndIf
    \EndFor    
\EndFor 
\State $B_1 \gets$ States2Boolean($V$); $B_0 \gets$ States2Boolean($\bar{V}$)
\State $\phi \gets$ Quine-McCluskey(ones=$B_1$, zeros=$B_0$) 
\State translate $\phi$ to explanations $\cE$ via language templates
\State \Return $\cE$
\Function{State2Boolean}{$W$}
\State $B \gets \{\}$
\ForAll{$\vb{s}=(s^1, \dots, s^N) \in W$}
    \For{$1 \le i \le N$ \blue{[replace with $i \in G$]}}
        \State $C \gets$ feature predicate valuations of state $s^i$
        \For{$f \in \cF$ \blue{[replace with $f \in F$]}}
            \State insert $C(f)$ to $B$
        \EndFor
    \EndFor
\EndFor
\State \Return $B$
\EndFunction
\end{algorithmic}
\end{algorithm}



\begin{algorithm}[h]
\footnotesize
\caption{Answering ``what'' query}
\label{ag:what}
\textbf{Input}: policy abstraction $\cM=(\cS, \cA, \cT)$, query ``what do agents $G_q$ do when satisfying predicates $F_q$''  \\
\textbf{Output}: language explanations $\cE$
\begin{algorithmic}[1] 
\State $A_q \gets \{\}$ 
\State \blue{$\alpha \gets$ find all relevant actions of predicates $F_q$}
\ForAll{joint state $\vb{s}=(s^1, \dots, s^N) \in \cS$}
    \If{$s^i$ satisfies $F_q$ for all $i \in G_q$}
        \ForAll{$\vb{a}=(a^1, \dots, a^N)$ enabled in $\vb{s}$}
            \ForAll{$i \in G_q$}
                \State insert $a^i$ to $A_q$ \blue{[only if $a^i \in \alpha$]}
            \EndFor
        \EndFor
    \EndIf
\EndFor
\State \blue{$A_q \gets$ the most frequent action for each agent in $A_q$}
\State generate explanations $\cE$ with $A_q$ via language templates
\State \textbf{return} $\cE$
\end{algorithmic}
\end{algorithm}






    


\newpage
\section{User Study Details} \label{app:study}

\startpara{Questionnaire on explanation goodness metrics}
Participants were instructed to rate on a 5-point Likert scale (1 - strongly disagree, 5 - strongly agree) about the following statements, which were adapted from~\cite{hoffman2018metrics}.
\begin{itemize}
    \item The explanations help me \emph{understand} how the team of robots completes the search and rescue mission.
    \item The explanations are \emph{satisfying}.
    \item The explanations are sufficiently \emph{detailed}.
    \item The explanations are sufficiently \emph{complete}, that is, they provide me with all the needed information to answer the questions.
    \item The explanations are \emph{actionable}, that is, they help me know how to answer the questions.
    \item The explanations let me know how \emph{reliable} the robot team is for completing the mission.
    \item The explanations let me know how \emph{trustworthy} the robot team is for completing the mission.
\end{itemize}

\startpara{User interface and sample questions}
Figures~\ref{fig:sample_query_when}-\ref{fig:sample_summary_gif} show examples of user interface and questions presented to participants during the user study.

\begin{figure}[h]
    \centering
    \includegraphics[width=1\columnwidth]{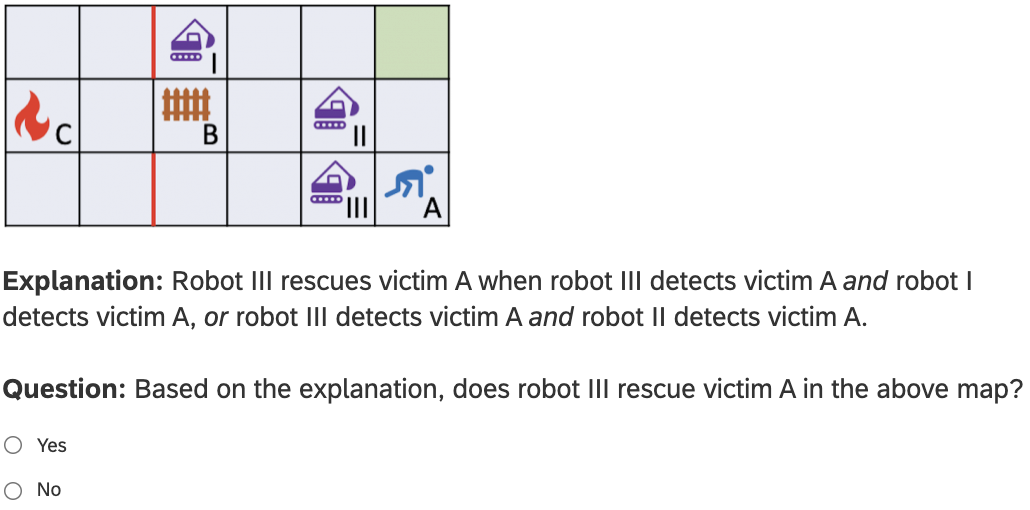}
    \caption{Question based on explanations for a ``when'' query.}
    \label{fig:sample_query_when}
\end{figure}

\begin{figure}[h]
    \centering
    \includegraphics[width=1\columnwidth]{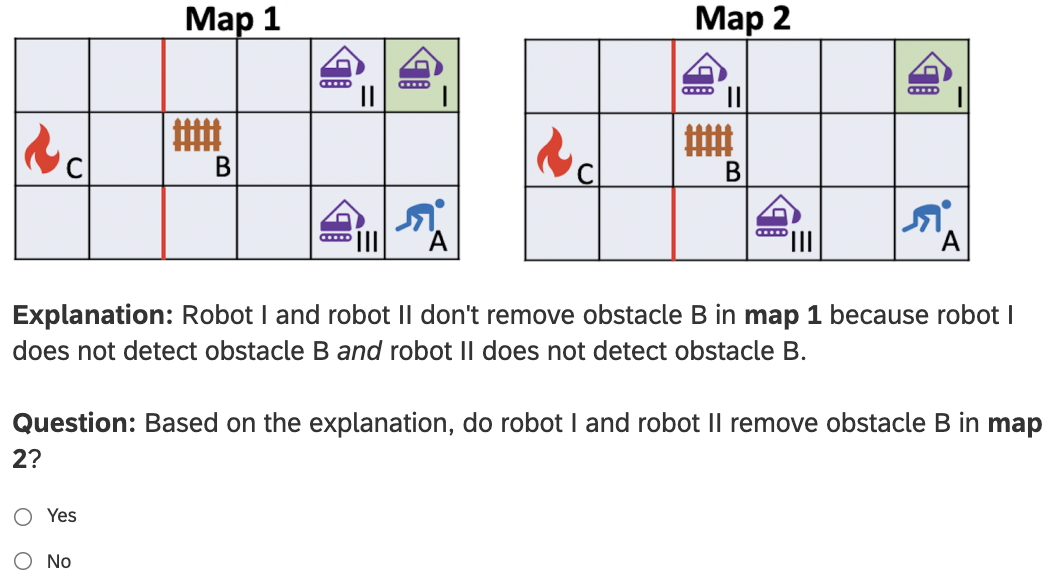}
    \caption{Question based on explanations for a ``why not'' query.}
    \label{fig:sample_query_whynot}
\end{figure}

\begin{figure}[h]
    \centering
    \includegraphics[width=1\columnwidth]{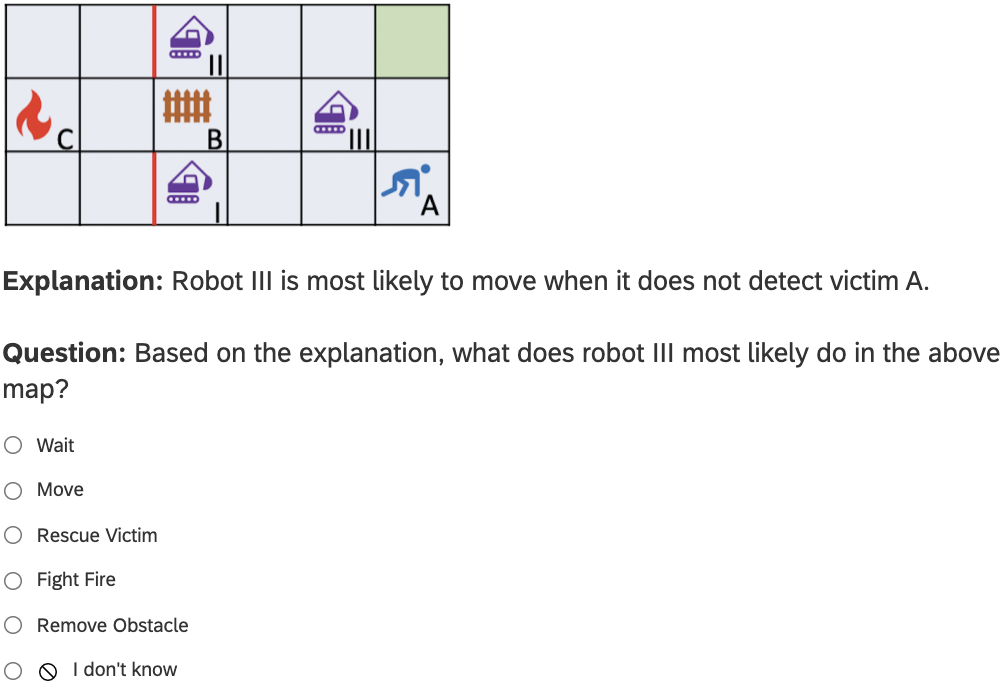}
    \caption{Question based on explanations for a ``what'' query.}
    \label{fig:sample_query_what}
\end{figure}

\begin{figure}[h]
    \centering
    \includegraphics[width=1\columnwidth]{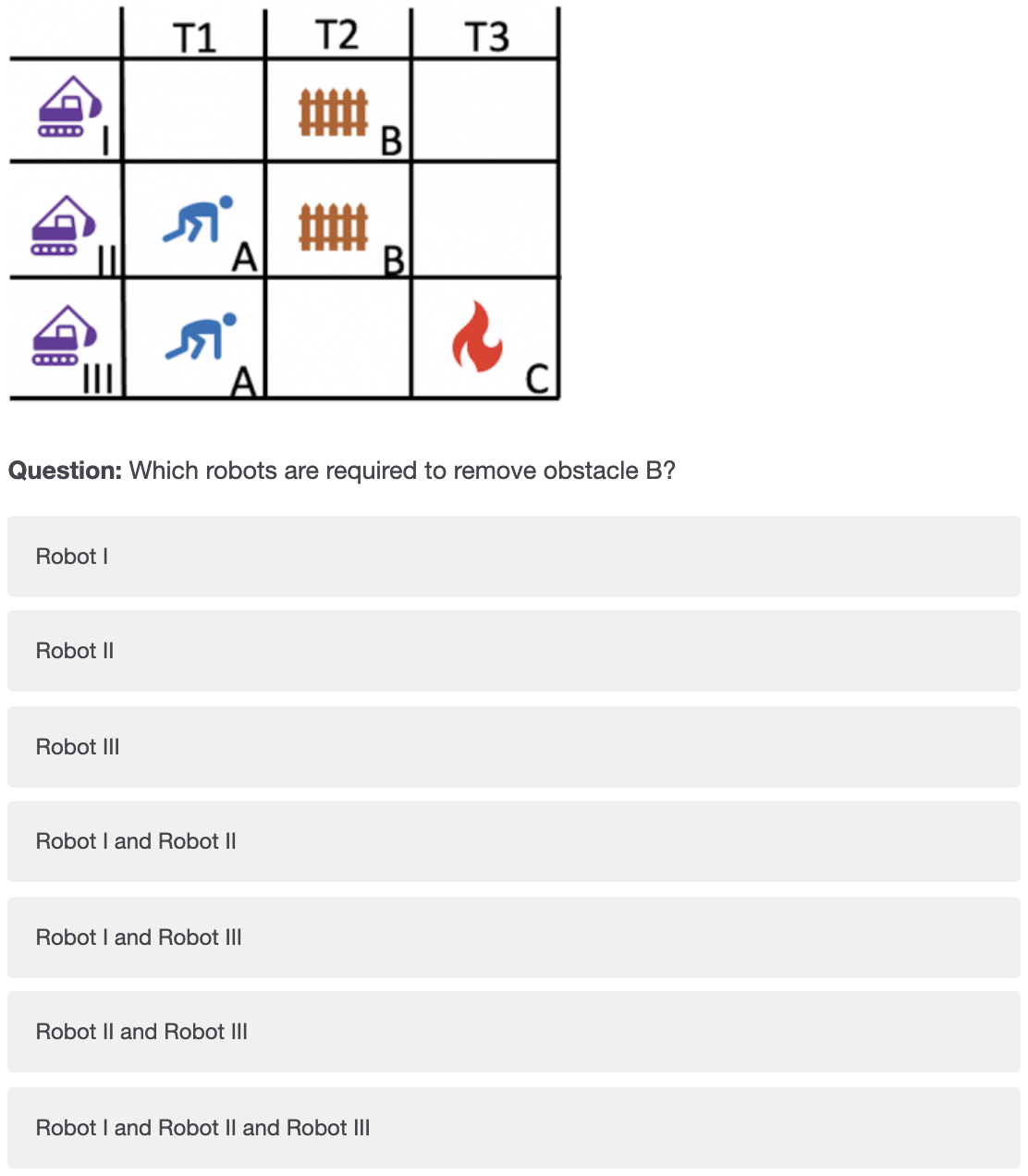}
    \caption{Question based on policy summarization (sequence chart).}
    \label{fig:sample_summary_schedule}
\end{figure}

\begin{figure}[h]
    \centering
    \includegraphics[width=1\columnwidth]{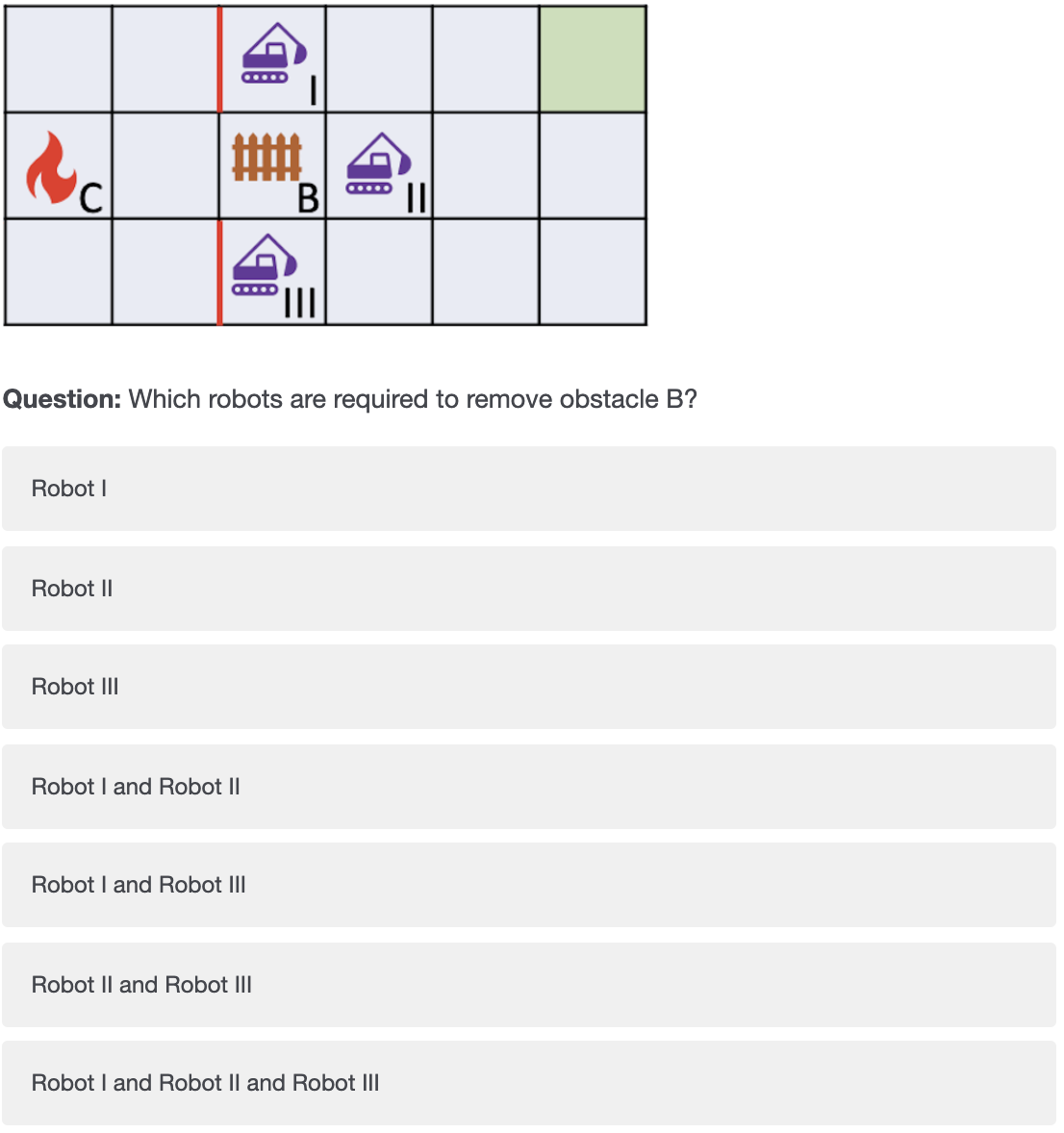}
    \caption{Question based on a policy summarization (GIF animation: https://github.com/kjboggess/IJCAI2022/blob/main/MissionGifExample.gif}
    \label{fig:sample_summary_gif}
\end{figure}

\fi

\end{document}